\documentclass[10pt, a4paper, twocolumn]{article} 

\usepackage[english]{babel} 

\usepackage{microtype} 

\usepackage{amsmath,amsfonts,amsthm} 

\usepackage[svgnames]{xcolor} 

\usepackage[hang, small, labelfont=bf, up, textfont=it]{caption} 

\usepackage{booktabs} 

\usepackage{lastpage} 

\usepackage{graphicx} 

\usepackage{siunitx}

\usepackage{enumitem} 
\setlist{noitemsep} 

\usepackage{float}

\usepackage[draft]{fixme}

\usepackage{sectsty} 
\allsectionsfont{\usefont{OT1}{phv}{b}{n}} 

\usepackage{geometry} 

\usepackage[T1]{fontenc} 
\usepackage[utf8]{inputenc} 

\usepackage{XCharter} 

\usepackage{fancyhdr} 
\fancypagestyle{firstpage}{ 
	\fancyhf{}
}

\newcommand{\authorstyle}[1]{{\large\usefont{OT1}{phv}{b}{n}\color{DarkRed}#1}} 

\newcommand{\institution}[1]{{\footnotesize\usefont{OT1}{phv}{m}{sl}\color{Black}#1}} 

\usepackage{titling} 

\newcommand{\HorRule}{\color{DarkGoldenrod}\rule{\linewidth}{1pt}} 

\pretitle{
	\vspace{-30pt} 
	\HorRule\vspace{10pt} 
	\fontsize{32}{36}\usefont{OT1}{phv}{b}{n}\selectfont 
	\color{DarkRed} 
}

\posttitle{\par\vskip 15pt} 

\preauthor{} 

\postauthor{ 
	\vspace{10pt} 
	\par\HorRule 
	\vspace{20pt} 
}

\usepackage{lettrine} 
\usepackage{fix-cm}	

\usepackage{xstring} 

\usepackage[backend=bibtex,style=numeric,natbib=true,doi=false,isbn=false,url=false]{biblatex} 

\usepackage[autostyle=true]{csquotes} 

\title{Spinal cord gray matter segmentation using deep dilated convolutions} 

\author{
	\authorstyle{Christian S. Perone\textsuperscript{1}, Evan Calabrese\textsuperscript{2,3}, Julien Cohen-Adad\textsuperscript{1,4}} 
	\newline\newline 
	\textsuperscript{1}\institution{Polytechnique Montreal, Montreal, QC, Canada}\\ 
	\textsuperscript{2}\institution{Department of Radiology, Center for in Vivo Microscopy, Duke University Medical Center, Durham, North Carolina, USA}\\
	\textsuperscript{3}\institution{Department of Radiology \& Biomedical Imaging, University of California San Francisco, San Francisco, California, USA}\\
	\textsuperscript{4}\institution{Functional Neuroimaging Unit, CRIUGM, Université de Montréal, Montreal, QC, Canada}\\ 
}

\date{\parbox{\linewidth}{\centering%
		\today\endgraf\bigskip
		Coordinator 1 \hspace*{3cm} Coordinator 2\endgraf\medskip
		Dept.\ of Physics \endgraf
		ABC College}}

\date{\today \\ 
\emph{* Article under submission to \textbf{Nature Scientific Reports}.}
} 

\begin{document}

\maketitle 

\thispagestyle{firstpage} 


\abstract{Gray matter (GM) tissue changes have been associated with a wide range of neurological disorders and was also recently found relevant as a biomarker for disability in amyotrophic lateral sclerosis. The ability to automatically segment the GM is, therefore, an important task for modern studies of the spinal cord. In this work, we devise a modern, simple and end-to-end fully automated human spinal cord gray matter segmentation method using Deep Learning, that works both on \emph{in vivo} and \emph{ex vivo} MRI acquisitions. We evaluate our method against six independently developed methods on a GM segmentation challenge and report state-of-the-art results in 8 out of 10 different evaluation metrics as well as major network parameter reduction when compared to the traditional medical imaging architectures such as U-Nets.}


\section{Introduction}
Gray matter (GM) and white matter (WM) tissue changes in the spinal cord (SC) have been linked to a large spectrum of neurological disorders \cite{Amukotuwa2007}. For example, using magnetic resonance imaging (MRI), the involvement of the spinal cord gray matter (SCGM) area in multiple sclerosis (MS) was found to be the strongest correlate of disability in multivariate models including brain GM and WM volumes, FLAIR lesion load, T1-lesion load, SCWM area, number of spinal cord T2 lesions, age, sex and disease duration \cite{Schlaeger2014a}. Another study showed that SCGM atrophy is a relevant biomarker for predicting disability in amyotrophic lateral sclerosis \cite{PaquinM-EElMendiliMMGrosCDupontSCohen-AdadJ2017}. 

The ability to automatically assess and characterize these changes is, therefore, an important required step \cite{DeLeener2016} in the modern pipeline to study both the \emph{in vivo} and \emph{ex vivo} SC. The segmentation outcome can also be used for co-registration and spatial normalization to a common space. Moreover, the fully-automated segmentation is very useful for longitudinal studies, where the delineation of gray matter is very time-consuming \cite{DeLeener2016}.

While recent cervical cord cross-sectional area (CSA) segmentation methods have achieved near-human performance \cite{DeLeener2014}, the accurate segmentation of the GM is still a remaining challenge \cite{Prados2017}. The main properties that make the GM area difficult to segment are: inconsistent surrounding tissue intensities, image artifacts and pathology-induced changes in the image contrast \cite{DeLeener2016}.

\begin{figure*}
	\centering
	\includegraphics[width=\textwidth]{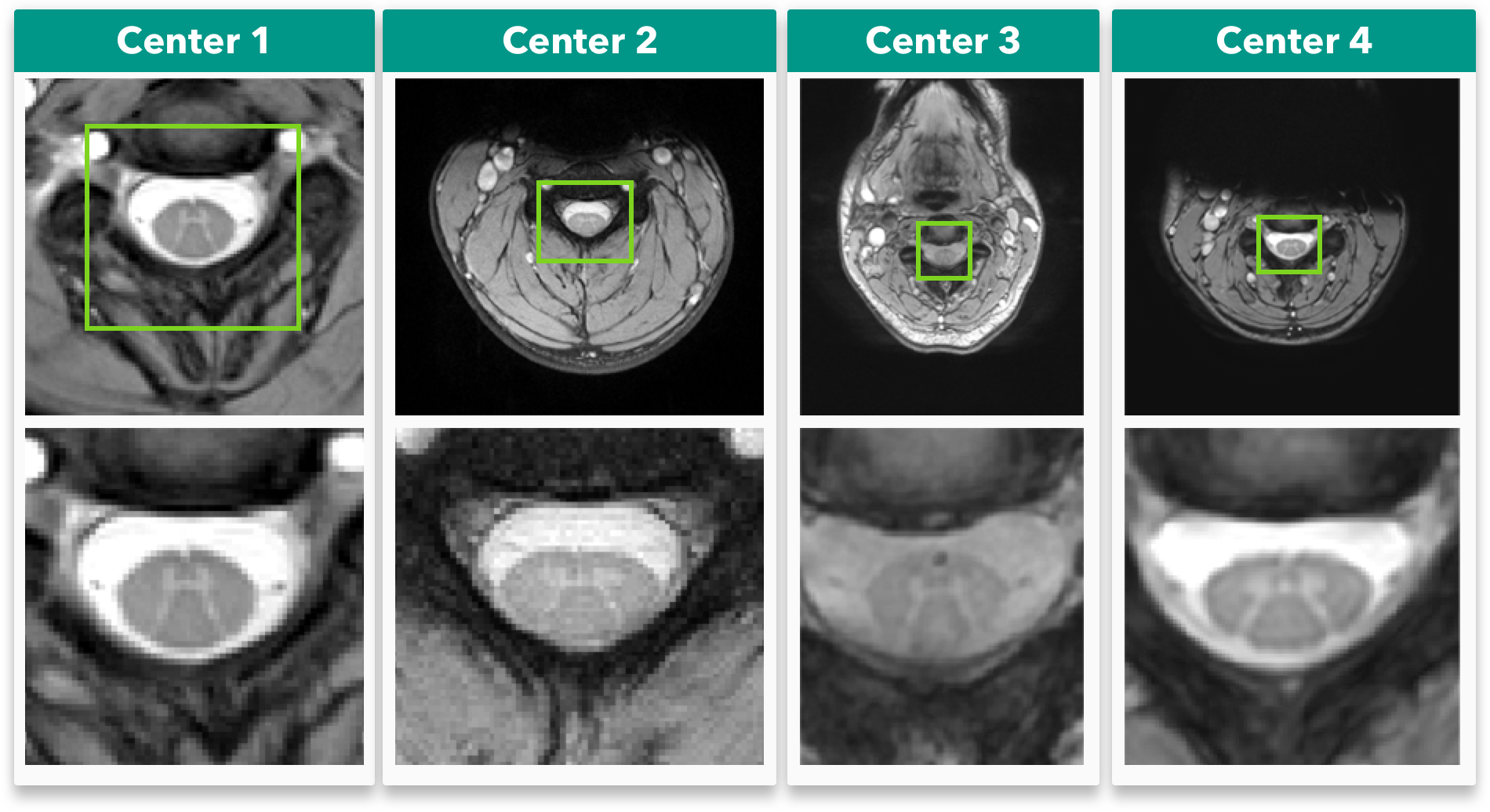}
	\caption{In vivo axial-slice samples from four centers (UCL, Montreal, Zurich, Vanderbilt) that collaborated to the SCGM Segmentation Challenge \cite{Prados2017}. Top row: original MRI images. Bottom row: crop of the spinal cord (green rectangle).}
	\label{fig:gmsample}
\end{figure*}

Other factors also contribute to the complexity of the GM segmentation task, such as lack of standardized data sets, differences in MRI acquisition protocols, different pixel sizes, different methods to acquire gold standard segmentations and different performance metrics to assess segmentation results \cite{Prados2017}. In Figure \ref{fig:gmsample}, we show some MRI samples (axial slices) acquired in different centers, where we can visually see the variability present in different acquisitions.

However, despite these difficulties, the scientific community recently organized a joint collaboration effort called "Spinal Cord Gray Matter Segmentation Challenge" (SCGM Challenge) \cite{Prados2017} to characterize the state-of-the-art and compare six independent developed methods \cite{Prados2016a}\cite{Brosch2016}\cite{Datta2017a}\cite{Dupont2017a}\cite{Blaiotta2016}\cite{Porisky2017} on a public available standard data set created through the collaboration of four internationally recognized spinal cord imaging research groups (University College London, Polytechnique Montreal, University of Zurich and Vanderbilt University), providing therefore a ground basis for method comparison that was previously unfeasible. 

In the past few years, we witnessed the fast and unprecedented development of Deep Learning \cite{LeCun2015a} methods, that not only achieved human-level performance but also surpassed it \cite{He2015b}, even in health domain applications \cite{Rajpurkar2017}. After the groundbreaking results presented in the seminal paper of the AlexNet \cite{Krizhevsky2012a}, the community embraced the successful Deep Learning approach for machine learning and consequently developed many methods that are nowadays state-of-the-art and pervasive in many different fields such as image classification \cite{He2016}, image segmentation \cite{Chen2017}, speech recognition \cite{Amodei2015}, natural language processing (NLP), among others.

Deep Learning is characterized by a major shift from the past traditional handcraft feature extraction to a hierarchical representation learning approach where multiple levels of automatically discovered representations are learned from raw data \cite{LeCun2015a}.

On a recent survey \cite{Litjens2017} that reviewed over 300 papers using Deep Learning techniques for medical image analysis, the authors found that Deep Learning techniques have spread throughout the entire field of medical image analysis, with a rapid increase in the number of published studies between the years of 2015 and 2016. The survey also found that Convolutional Neural Networks (CNNs) were more prevalent in the medical image analysis, with Recurrent Neural Networks (RNNs) gaining more popularity.

Although the enormous success of Deep Learning has attracted a lot of attention of the research community, some challenges in the medical imaging domain still remain open:
\begin{itemize}
	\item \textbf{Data acquisition} is usually very expensive and require time-consuming specialist annotation to create gold standards;  
	\item \textbf{Standardized data sets} are still a major problem due to variability in equipment from different vendors, acquisition protocols/parameters/contrasts, especially in the MRI domain;
	\item \textbf{Data availability} is also limited due to privacy/ethics concerns or regulations \cite{Litjens2017};	
\end{itemize}

In this work, we propose a new simple pipeline with an end-to-end learning approach for fully automated spinal cord gray matter segmentation using a novel Deep Learning architecture based on the \emph{Atrous} Spatial Pyramid Pooling (ASPP) \cite{Chen2016c}\cite{Chen2017}, where we achieved state-of-the-art results on many metrics in an \emph{in vivo} independent data set evaluation. We also show excellent generalization on an \emph{ex vivo} high-resolution acquisition data set where only a few axial-slices were annotated to accurate segment an MRI volume with more than 4000 axial slices.

We also provide an evaluation comparing our method with the traditionally used U-Net \cite{Ronneberger2015} architecture and with other six independently developed methods. 

\section{Related Work}
Many methods for spinal cord segmentation were proposed in the past. Regarding the presence or absence of manual intervention, the segmentation methods can be separated in two main categories: semi-automated and fully-automated. In \cite{DeLeener2016}, the authors also classify spinal cord segmentation methods in the following categories:
\begin{itemize}
	\item \textbf{Intensity-based}: examples are threshold-based or edge-detection methods;
	\item \textbf{Surface-based}: examples include deformable models, active-contours and level-set methods;
	\item \textbf{Image-based}: examples include watershed and template/atlas deformation methods;
\end{itemize}

In \cite{Blaiotta2016}, they propose a probabilistic method for segmentation called "Semi-supervised VBEM", where the MRI signals are assumed to be observed data generated by warping of an average shaped reference anatomy \cite{Prados2017}. The observed image intensities are modeled as random variables drawn from a Gaussian mixture distribution, where the parameters are estimated using a variational version of the Expectation-Maximization (EM) \cite{Blaiotta2016} algorithm. The method can be used in a fully unsupervised fashion or by incorporating training data with manual labels, hence the semi-supervised scheme \cite{Prados2017}. 

The SCT (Spinal Cord Toolbox) segmentation method \cite{Dupont2017a}, uses an atlas-based approach and was built based on a previous work \cite{Asman2014a} but with additional improvements such as the use of vertebral level information and linear intensity normalization to accommodate multi-site data \cite{Dupont2017a}. The SCT approach first builds a dictionary of images using manual WM/GM segmentations after a pre-processing step, then the target image is also pre-processed and normalized, after that, the target image is projected into the PCA (Principal Component Analysis) space of the dictionary images where the most similar dictionary slices are selected using an arbitrary threshold, and finally, the segmentation is done using label fusion between the manual segmentations from the dictionary images that were selected \cite{Prados2017}. The SCT method is freely available as an open-source software\footnote{https://github.com/neuropoly/spinalcordtoolbox} package \cite{DeLeener2017}.

In \cite{Prados2016a}, they propose a method called "Joint collaboration for spinal cord gray matter segmentation" (JCSCS), where two existing label fusion segmentation methods were combined. The method is based on a multi-atlas segmentation propagation using registration and segmentation in 2D slice-wise space. In JCSCS, the "Optimized PatchMatch Label Fusion" (OPAL) \cite{Giraud2016} is used to detect the spinal cord, where the cord localization is achieved by providing an external data set of spinal cord volumes and their associated manual segmentation \cite{Prados2016a}, after that, the "Similarity and Truth Estimation for Propagated Segmentations" (STEPS) \cite{JorgeCardoso2013} is used to segment the GM in two steps, first the segmentation propagation, and then a consensus segmentation is created by fusing best-deformed templates (based on locally normalized cross-correlation) \cite{Prados2016a}.

In \cite{Datta2017a}, the Morphological Geodesic Active Contour (MGAC) algorithm uses an external spinal cord segmentation tool (Jim, from Xinapse Systems) to estimate the spinal cord boundary as well as a morphological geodesic active contour model to segment the gray matter. The method has five steps: first, the original image spinal cord is segmented with the Jim software and then a template is registered to the subject cord, after that the same transformation is applied to the GM template. The transformed gray matter template is then used as an initial guess for the active contour algorithm \cite{Datta2017a}.

The "Gray matter Segmentation Based on Maximum Entropy" (GSBME) algorithm \cite{Prados2017} is a semi-automatic, supervised segmentation method for the GM. The GSBME is comprised of three main stages. First, the image is pre-processed, in this step the GSBME uses the SCT \cite{DeLeener2017} to segment the spinal cord using Propseg \cite{DeLeener2014} with manual initialization, after that the intensities are normalized and denoised. In the second step, the images are slice-wise thresholded using a sliding window where the optimal threshold is found by maximizing the sum of the GM and WM intensity entropies. In the third and last stage, an outlier detector discards segmented intensities using morphological features such as perimeter, eccentricity and Hu moments among others \cite{Prados2017}.

In the Deepseg approach \cite{Porisky2017}, built on top of \cite{Brosch2016}, they use a Deep Learning architecture similar to the U-Net \cite{Ronneberger2015}, where a CNN has a contracting and expanding path. The contracting path aggregates information while the expanding path upsamples the feature maps in order to achieve a dense prediction output. To recover spatial information loss, shortcuts are added between contracting/expanding paths of the network. In Deepseg, instead of using upsampling layers like in U-Net, they use an unpooling and "deconvolution" approach such as in \cite{Zeiler2011}. The network architecture used has 11 layers and is pre-trained using 3 convolutional restricted Boltzmann Machines \cite{Lee2009}. Deepseg also uses loss function with a weighted sum of two different terms, the mean square differences of the GM and non-GM voxels, balancing sensitivity and specificity \cite{Prados2017}. Two models were trained (independently), one for the full spinal cord segmentation and another for the GM segmentation.

We compare our method with all the aforementioned methods on the SCGM Challenge \cite{Prados2017} data set.

\section{Methods and Materials}
As we saw in the \emph{Related Work} section, the majority of the previously developed GM segmentation methods usually rely on registered templates/atlases, arbitrary distance and similarity metrics or complex pipelines that aren't optimized in an end-to-end fashion, neither efficient during inference time. 

In this work, we focus on the development of a simple Deep Learning method that can be trained in an end-to-end fashion and that generalizes well even with a small number of 2D labeled axial slices of a 3D MRI volume.

\subsection{Note on U-Nets}
Many modern Deep Learning CNN classification architectures use alternating layers of convolutions and subsampling operations to aggregate semantic information and discard spatial information across the network, leading to certain levels of translation and rotation invariance that are important for classification. However, in segmentation tasks, a dense full-resolution output is required. In medical imaging, the most traditional architecture for segmentation is the well-known U-Net \cite{Ronneberger2015}, where two distinct paths (encoder-decoder/contracting-expanding) are used to aggregate semantic information and recover the spatial information with the help of shortcut connections between the paths. 

The U-Net architecture, however, causes a major expansion of the parameter space due to the two distinct paths that form the U-shape. We also found, such as in \cite{Zhang}, that the gradient flow in the high-level layers of the U-Nets (bottom of the U-shape) is problematic. Since the final low-level layers have access to the earlier low-level features, the network optimization will find the shortest path to minimize the loss, thus reducing the gradient flow in the bottom of the network.

By visualizing feature maps from the U-Net using techniques described in \cite{Yosinski2015}, we found that the features extracted in the bottom of the network were very noisy while the features extracted in the low-level layers were the only ones showing meaningful patterns. By removing the bottom layers of the network, we found that the network performed the same or sometimes better than the deeper network.

\subsection{Proposed method}
Our method is based on the state-of-the-art segmentation architecture called "\emph{Atrous} Spatial Pyramid Pooling" (ASPP) \cite{Chen2017} that uses "\emph{Atrous} convolutions", also called "dilated convolutions" \cite{Yu2016}. We performed modifications to improve the segmentation performance on medical imaging by handling imbalanced data with a different loss function and also by extensively removing decimation operations from the network such as pooling, trading depth (due to memory constraints) to improve the translational equivariance of the network and also parameter reduction. 

\begin{figure}[H]
	\includegraphics[width=\linewidth]{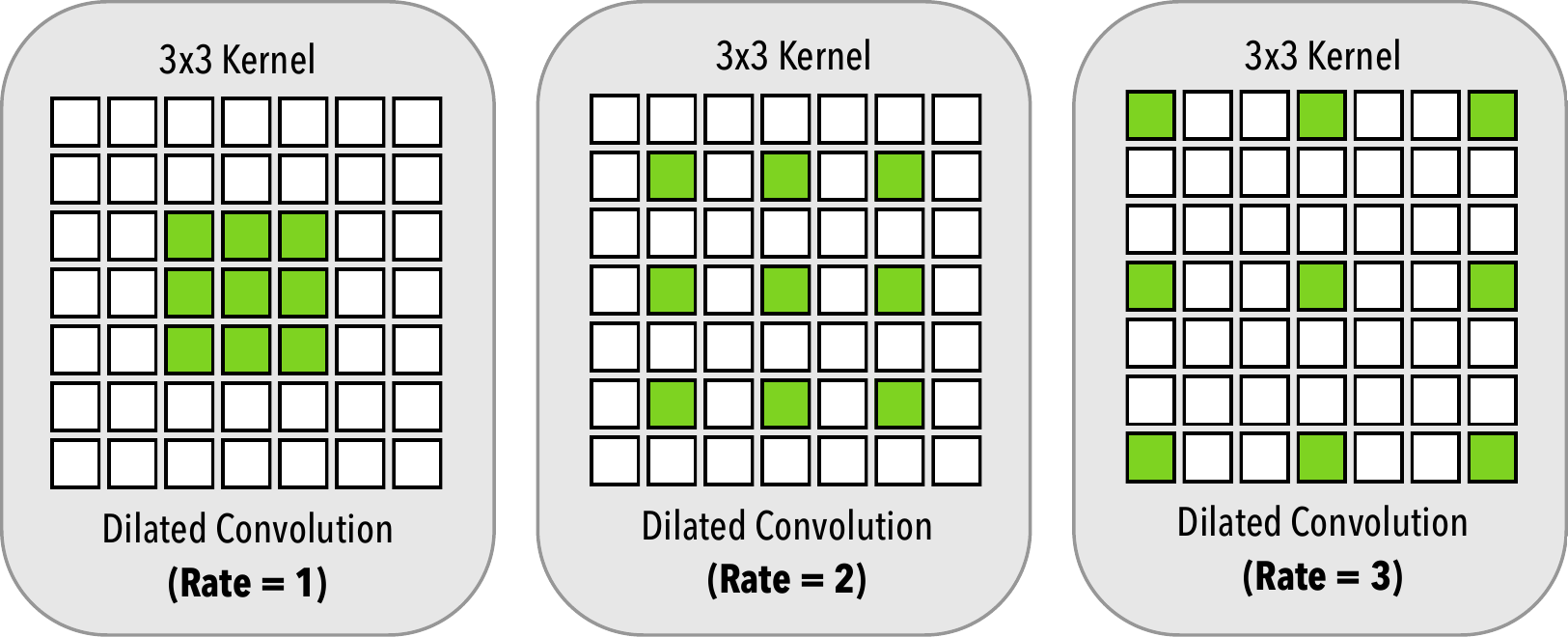}
	\caption{Dilated convolution. On the left, we have the dilated convolution with dilation rate $r = 1$, equivalent to the standard convolution. In the middle with have a dilation $r = 2$ and in the right a dilation rate of $r = 3$. All dilated convolutions have a 3x3 kernel size and the same number of parameters.}
	\label{fig:dilated-conv}
\end{figure}

Dilated convolutions allow us to exponentially grow the receptive field with linearly increasing number of parameters, providing a significant parameter reduction while increasing the effective receptive field. Dilated convolutions works by introducing "holes" \cite{Chen2016c} in the kernel as illustrated in Figure \ref{fig:dilated-conv}. For an 1D signal $x[i]$, the $y[i]$ output of a dilated convolution with the dilation rate $r$ and a filter $w[s]$ with size $S$ is formulated as:

\begin{align}
y[i] = \sum_{s=1}^S x[i + r \cdot s] w[s].
\end{align}

The dilation rate $r$ can also be seen as the stride to which the input signal is sampled \cite{Chen2016c}. Dilated convolutions, like standard convolutions, also have the advantage of being translationally equivalent, which means that translating the image will result in a translated version of the original input, as seen below:

\begin{align}
	f(g(x)) = g(f(x))
\end{align}

Where $g(\cdot)$ is a translation operation and $f(\cdot)$ a convolution operation. However, since we don't need to introduce pooling to capture multi-scale features when using dilated convolutions, we can keep the translational equivariance property in the network, which is very important for spatial dense prediction tasks.

\begin{figure*}
	\centering
	\includegraphics[width=\textwidth]{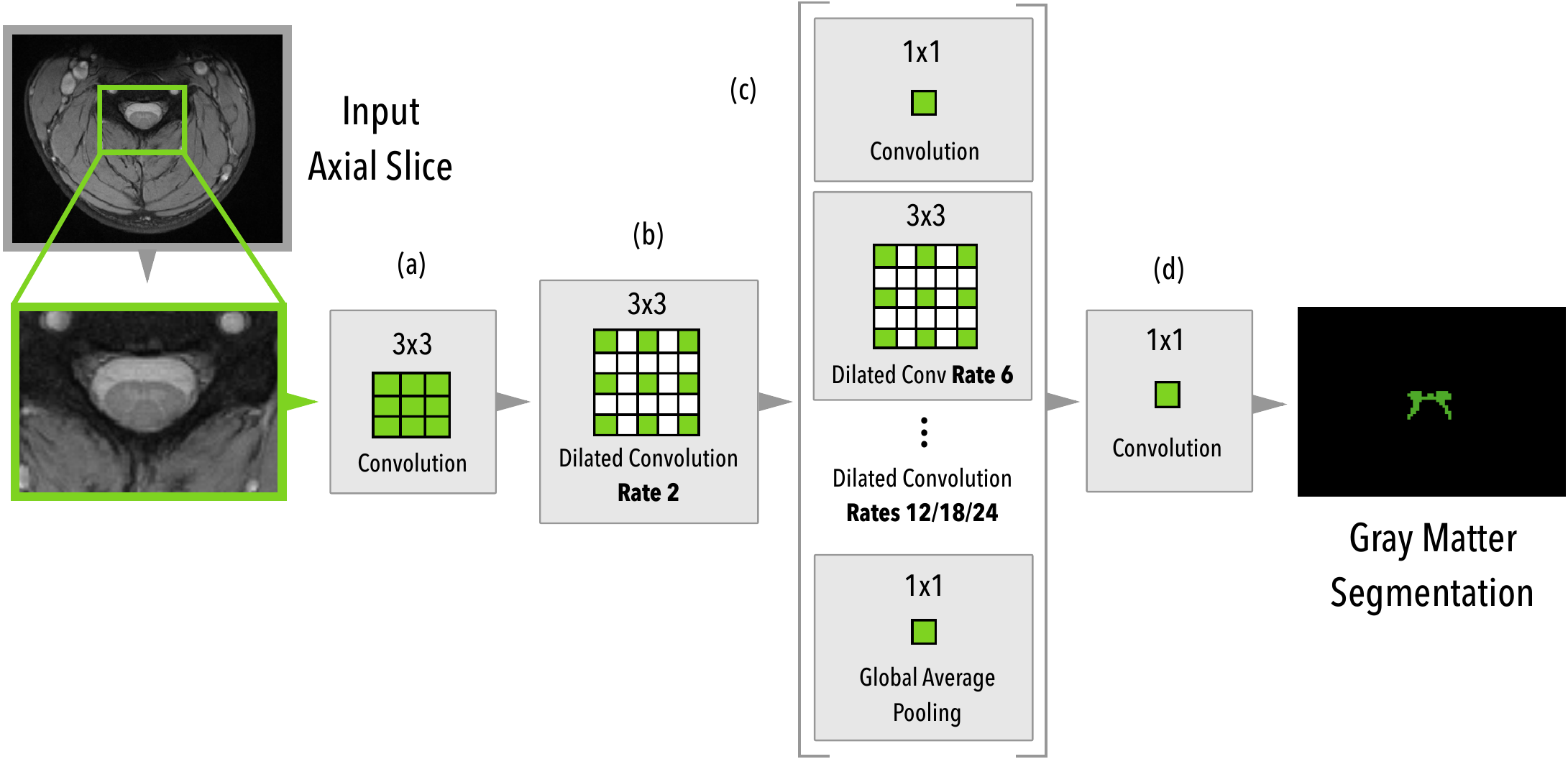}
	\caption{Architecture overview of the proposed method. The MRI axial slice is fed to the first block of 3x3 convolutions and then to a block of dilated convolutions (rate 2). Then, six parallel modules with different rates (6/,12/18/24), 1x1 convolution, and a global average pooling are used. After the parallel modules, all feature maps are concatenated and then fed into the final block of 1x1 convolutions to produce the final dense predictions.}
	\label{fig:architecture}
\end{figure*}

The overall proposed architecture can be seen in Figure \ref{fig:architecture}. Our architecture works with 2D slice-wise axial images and is composed of (a) two initial layers of standard 3x3 convolutions, followed by (b) two layers of dilated convolutions with rate $r = 2$, followed by (c) six parallel branches with two layers each of a 1x1 standard convolution, 4 different dilated convolution rates (6/12/18/24) and a global averaging pooling that is repeated at every spatial position of the feature map. After that, the feature maps from the six parallel branches are concatenated and forwarded to (d) a block of 2 layers with 1x1 convolutions in order to produce the final dense prediction probability map. Each layer is followed by Batch Normalization \cite{Ioffe2015} and Dropout \cite{Srivastava2014} layers. 

\begin{figure}
	\centering
	\includegraphics[width=\linewidth]{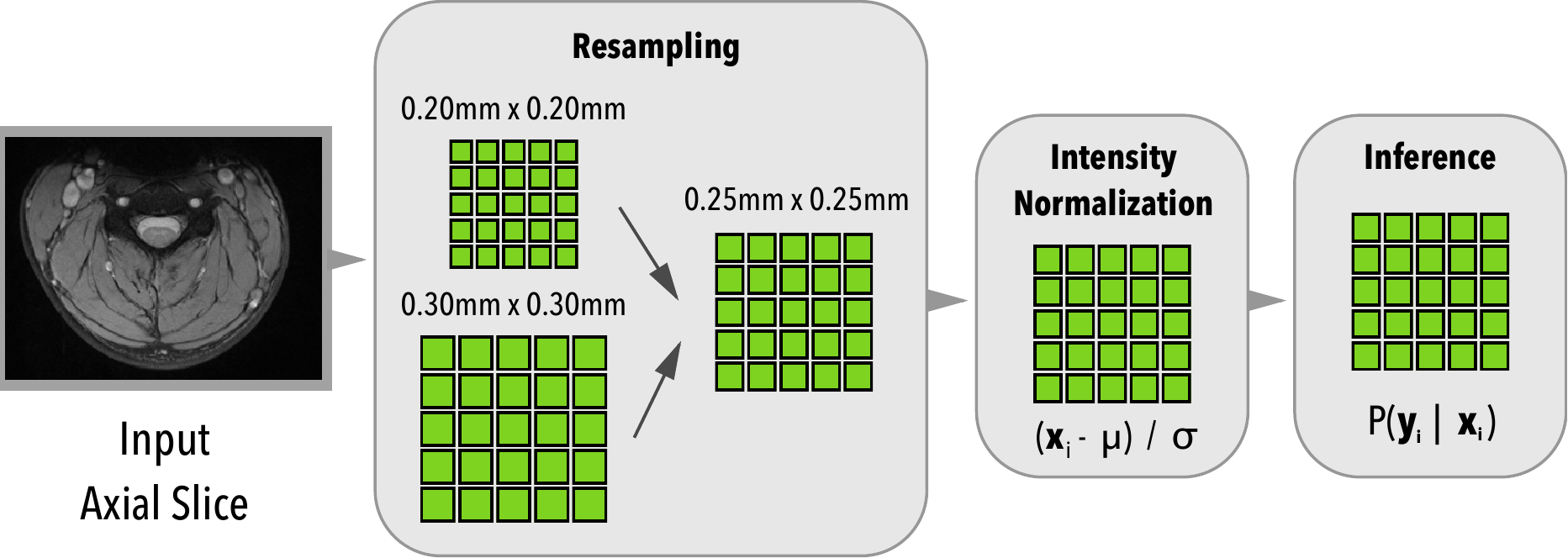}
	\caption{Architecture pipeline overview. During the first stage, input axial slices are resampled to a common pixel size space, then intensity is normalized, followed by the network inference.}
	\label{fig:arch-pipeline}
\end{figure}

Figure \ref{fig:arch-pipeline} illustrates the pipeline of our training/inference process. An initial resampling step downsamples/upsamples the input axial slice images to a common pixel size space, then a simple intensity normalization is applied to the image, followed by the network inference stage.

Contrary to the task of natural images segmentation, the task of GM segmentation in medical imaging is usually very unbalanced. In our case, only a small portion of the entire axial slice encompasses the GM (the rest being comprised of other structures such as the white matter, cerebrospinal fluid, bones, muscles, etc.). Due to this imbalance, we employed a surrogate loss for the DSC (Dice Similarity Coefficient) called the Dice Loss, which is insensitive to imbalancing and was employed by many works in medical imaging \cite{Milletari2016}\cite{Drozdzal2017}. The Dice Loss can be formulated as:

\begin{align}
 \mathcal{L}_{dice}= - \dfrac{2\sum_{n=1}^{N}p_{n}r_{n}+\epsilon}{\sum_{n=1}^{N}p_n + \sum_{n=1}^{N} r_n + \epsilon}
\end{align}

Where $p$ and $r$ are the predictions and gold standard respectively. The $\epsilon$ term is used to ensure the loss stability by avoiding the numerical issues. We experimentally found that the Dice Loss yielded better results when compared to the weighted cross-entropy (WCE) used by \cite{Ronneberger2015}, which is more difficult to optimize due to the added weighting hyper-parameter.

Medical image data sets are usually smaller than natural image data sets by many orders of magnitude, therefore regularization and data augmentation is an important step. In this work, the following data augmentation strategies were applied: rotation, shifting, scaling, flipping, noise and elastic deformation.

The main differences when we compare our architecture with \cite{Chen2017}, are the following:
\begin{description}
	\item[Initial pooling/decimation:] our network does not use initial pooling layers as we found them detrimental to the segmentation of medical images;
	\item[Padding:] we extensively employ padding across the entire network to keep feature map sizes fixed, trading depth to reduce memory usage of the network;
	\item[Dilation Rates:] since we don't use initial pooling, we kept the parallel dilated convolution branch with rate $r = 24$, as we found improvements by doing so, due to the large feature map size that doesn't cause filter degeneration as seen in \cite{Chen2017};
	\item[Loss:] contrary to natural images, our task of GM segmentation is very unbalanced, instead of traditional cross-entropy, we used the Dice Loss;
	\item[Data Augmentation:] in this work we applied not only scaling and flipping as seen in \cite{Chen2017} but also rotation, shifting, added channel noise and elastic deformations \cite{Simard2003}.
\end{description}

Table \ref{tab:method-comparison} compares the setup parameters of our approach as well as the methods that participated in the SCGM Segmentation Challenge \cite{Prados2017}.

\subsubsection{U-Net architecture}
For the U-Net \cite{Ronneberger2015} architecture model that was used for comparison, we employed a 14-layers network using standard 3x3 2D convolution filters with ReLU non-linearity activations. For a fair comparison, we used the same training protocol and loss function. For the data augmentation strategy, we employed a more aggressive augmentation due to overfitting issues with the U-Net that we'll discuss later.
We also performed a extensive architecture exploration, and used the best performing U-Net model architecture.

\begin{table}[]
	\centering
	\caption{Parameters of each compared method. Values replicated from \cite{Prados2017}. Time per slice are estimated values, since different hardware were employed by the different techniques.}
	\label{tab:method-comparison}
	\resizebox{\linewidth}{!}{
		\begin{tabular}{@{}lllll@{}}
			\toprule
			\textbf{Method}          & \textbf{Init.} & \textbf{Training}     & \textbf{External data} & \textbf{Time p/ slice} \\ \midrule
			\textbf{JCSCS}           & Auto.      & No                    & Yes                         & 4-5 min                 \\
			\textbf{DEEPSEG}         & Auto.      & Yes (4 h)             & No                          & \textless 1 s           \\
			\textbf{MGAC}            & Auto.      & No                    & No                          & 1 s                     \\
			\textbf{GSBME}           & Manual         & Yes (\textless 1 m) & No                          & 5-80 s                  \\
			\textbf{SCT}             & Auto.      & No                    & Yes                         & 8-10 s                  \\
			\textbf{VBEM}            & Auto.      & No                    & No                          & 5 s                     \\
			\textbf{Proposed} & Auto.      & Yes (19 h)            & No                          & \textless 1 s           \\ \bottomrule
		\end{tabular}
	}
\end{table}

\subsection{Data sets}
In this subsection, we present the data sets used for evaluation in this work.

\subsubsection{Spinal Cord Gray Matter Challenge}
The Spinal Cord Gray Matter Challenge \cite{Prados2017} (SCGM Challenge) data set is comprised by 80 healthy subjects (20 subjects from each center). The demographics ranges from a mean age of 28.3 up to 44.3 years old. Three different MRI systems were used (Philips Achieva, Siemens Trio, Siemens Skyra) with different acquisition parameters. The voxel size resolution of the data set ranges from 0.25x0.25x2.5 mm up to 0.5x0.5x5.0 mm. The data set is split between training (40) and test (40) with the test set hidden. For each labeled slice in the data set, 4 gold standard segmentation masks were produced by 4 independent expert raters (one per site).

Examples of the data set for each center are shown in the Figure \ref{fig:gmsample}.

During the development of this work, we found some misclassified voxels in the training set, these issues were reported, however for the sake of fair comparison, all the evaluations done in this work used the original pristine training data set.

\subsubsection{\textit{Ex vivo} high-resolution spinal cord}
To evaluate our method on an \textit{ex vivo} data set, we used an MRI acquisition that was performed on an entire human spinal cord, from the pyramidal decussation to the \textit{cauda equina} using a 7T horizontal-bore small animal MRI system.

Although the acquisition was obtained from a deceased adult male with no known history of neurologic disease, the review of images revealed a clinically occult SC lesion close to the 6th thoracic nerve root level, with imaging features suggestive of a chronic compressive myelopathy or possible sequela of a previous viral infection such as herpes zoster.

The volume is comprised by a total of 4676 axial slices with \SI{100}{\micro\metre} isotropic resolution and the acquisition time took approximately 120 hours.

\subsection{Training Protocol}

\subsubsection{Spinal Cord Gray Matter Challenge}
In this subsection we show the training protocol for the SCGM Challenge \cite{Prados2017} data set experiments.

\begin{description}
	\item[Resampling and Cropping:] All volumes were resampled to a voxel size of 0.25x0.25 mm, the highest resolution found between acquisitions. All the axial slices were center-cropped to a 200x200 pixels size;
	\item[Normalization:] We performed only mean centering and standard deviation normalization of the volume intensities;
	\item[Train/validation split:] For the train/validation split, we used 8 subjects (2 from each site) for validation and the rest for training. The test set was defined by the challenge. We haven't employed any external data or used the vertebral information from the provided dataset. Only the provided GM masks were used for training/validation;
	\item[Batch size:] We used a small batch size of only 11 samples;
	\item[Optimization:] We used Adam \cite{Kingma2015a} optimizer with a small learning rate $\eta = 0.001$;
	\item[Batch Normalization:] We used a momentum $\phi = 0.1$ for BatchNorm due to the small batch size;
	\item[Dropout:] We used a dropout rate of 0.4;
	\item[Learning Rate Scheduling:] Similar to \cite{Chen2017}, we used the "poly" learning rate policy where the learning rate is defined by:
	\begin{align}
		\eta = \eta_{t_{0}} * (1 - \frac{n}{N})^p
	\end{align}
	Where $\eta_{t_{0}}$ is the initial learning rate, $N$ is the number of epochs, $n$ the current epoch and $p$ the power with $p = 0.9$;
	\item[Iterations:] We trained the model for 1000 epochs (w/ 32 batches at each epoch);
	\item[Data augmentation:] We applied the following data augmentations: rotation, shift, scaling, channel shift, flipping and elastic deformation \cite{Simard2003}. The data augmentation parameters were chosen using random search;
\end{description}

Contrary to the very smooth decision boundaries that models trained using the traditional cross-entropy present, the Dice Loss has the property of creating very sharp decision boundaries and models with high recall rate. We found experimentally that thresholding the dense predictions with a threshold $\tau = 0.999$ provided a good compromise between precision/recall, however no optimization was employed to choose the threshold $\tau$ value for the output predictions.

Since the test data set is hidden from the challenge participants, to evaluate our model we sent our produced test predictions to the challenge website\footnote{http://cmictig.cs.ucl.ac.uk/niftyweb}. Results are presented in Table \ref{tab:challenge-results} on the column "Proposed Method", along with the other six previously developed methods and 10 different metrics.

The training time on a single NVIDIA P100 GPU took approximately\footnote{\label{notetf}Using single-precision floating-point (fp32) and TensorFlow 1.3.0 framework with cuDNN 6} 19 hours. While inference time took less than 1 second per subject.

\subsubsection{Inter-rater variability as label smoothing regularization}
The training data set provided by the SCGM Challenge is comprised of 4 different masks that were manually and independently created by raters for each axial slice. As in \cite{Brosch2016}, we used all the different masks as our gold standard. We also found that this approach shares the same principle of using label smoothing as seen in \cite{Szegedy2015}.

Label smoothing is a mechanism that make the model be less confident by preventing the network from assigning a full probability to a single class, usually an evidence of overfitting. In \cite{Pereyra2017}, they also found a link between label smoothing and confidence penalty through the direction of the Kullback–Leibler divergence.

Since the different gold standard masks for the same axial slices diverges usually only in the border of the GM, it is easy to see that this has a label smoothing effect on the contour of the GM, encouraging the model to be less confident in the contour prediction of the GM, a kind of ``contour smoothing''.

This interpretation suggests that one could also incorporate this contour smoothing by artificially adding label smoothing on the contours of the target anatomy, where raters usually disagree on the manual segmentation, leading to potentially better model generalization on many different medical segmentation tasks where the contours are usually the region of raters disagreement.

We leave the exploration of this contour smoothing to future work.

\subsubsection{\textit{Ex vivo} high-resolution spinal cord}
In this subsection we show the training protocol for the \textit{ex vivo} high-resolution spinal cord data set.

\begin{description}
	\item[Cropping:] All the slices were center-cropped to a 200x200 pixels size;
	\item[Normalization:] We performed only mean centering and standard deviation normalization of the volume intensities;
	\item[Train/validation split:] For the training set we selected only 15 evenly spaced axial slices out of 4676 total slices from the volume.  For the validation set, we selected 7 (evenly spaced) axial slices and our test set was comprised of 8 axial slices (also evenly distributed across the entire volume);
	\item[Batch size:] We used a small batch size of only 11 samples;
	\item[Optimization:] We used Adam \cite{Kingma2015a} optimizer with a small learning rate $\eta = 0.001$;
	\item[Dropout:] We used a dropout rate of 0.4;
	\item[Learning Rate Scheduling:] Similar to \cite{Chen2017}, we used the "poly" learning rate policy where the learning rate is defined by:
	\begin{align}
		\eta = \eta_{t_{0}} * (1 - \frac{n}{N})^p
	\end{align}
	Where $\eta_{t_{0}}$ is the initial learning rate, $N$ is the number of epochs, $n$ the current epoch and $p$ the power with $p = 0.9$;
	\item[Iterations:] We trained the model for 600 epochs (w/ 32 batches at each epoch);
	\item[Data augmentation:] For this dataset, we used the following aforementioned augmentations: rotation, shift, scaling, channel shift, flipping and elastic deformation \cite{Simard2003}. We didn't employed random search to avoid overfitting due to the data set size;
\end{description}

Like in the SCGM Segmentation task, we used a threshold $\tau = 0.999$ to binarize the prediction mask.

The training time on a single NVIDIA P100 GPU took approximately\footnote{See footnote \ref{notetf}.} 2 hours. While inference time took approximately 25 seconds to segment 4676 axial slices.

\section{Results}
In this section, we discuss the experimental evaluation of the method in the presented data sets.

\subsection{Spinal Cord Gray Matter Challenge}
In this subsection we show the evaluation on the SCGM Challenge \cite{Prados2017} data set.

\subsubsection{Qualitative Evaluation}
\begin{figure*}
	\centering
	\includegraphics[width=\textwidth]{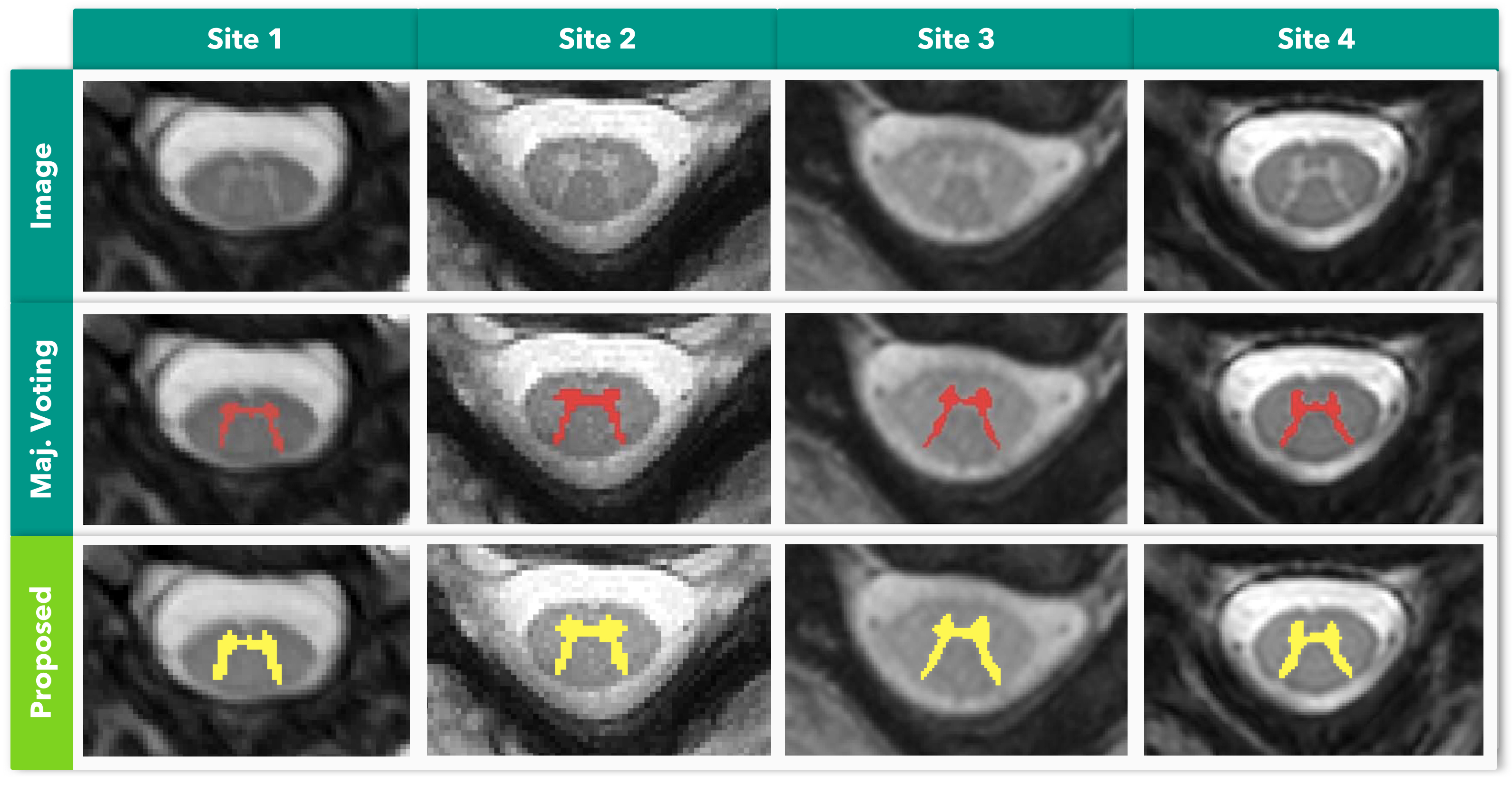}
	\caption{Qualitative evaluation of our proposed approach on the same axial slice for subject 11 of each site. From top to bottom row: input image, majority voting segmentation gold standard and the result of our segmentation method. Adapted from \cite{Prados2017}.}
	\label{fig:challenge-seg}
\end{figure*}

In Figure \ref{fig:challenge-seg}, we show the segmentation output of our model in four different subjects from acquisitions of four different centers on the test set of the SCGM Segmentation Challenge. The majority voting segmentation was taken from \cite{Prados2017}. As we can see in Figure \ref{fig:challenge-seg}, our approach was able to capture many properties of the GM anatomy, providing good segmentations even in presence of blur as seen in the samples from the Site 1 and Site 3.

When compared with the segmentation results from Deepseg \cite{Porisky2017}, that uses a U-Net like structure with pre-training and 3D-wise training, we can see that our method doesn't fail to segment the gray commissure of the GM structure as seen in the Figure 4 of \cite{Prados2017}.

\subsubsection{Quantitative Evaluation}
As we can see in Table \ref{tab:challenge-results}, our approach achieved state-of-the-art results in 8 out of 10 different metrics and surpassed 4 out of 6 previously developed methods on all metrics.

\begin{figure*}
	\centering
	\includegraphics[width=\textwidth]{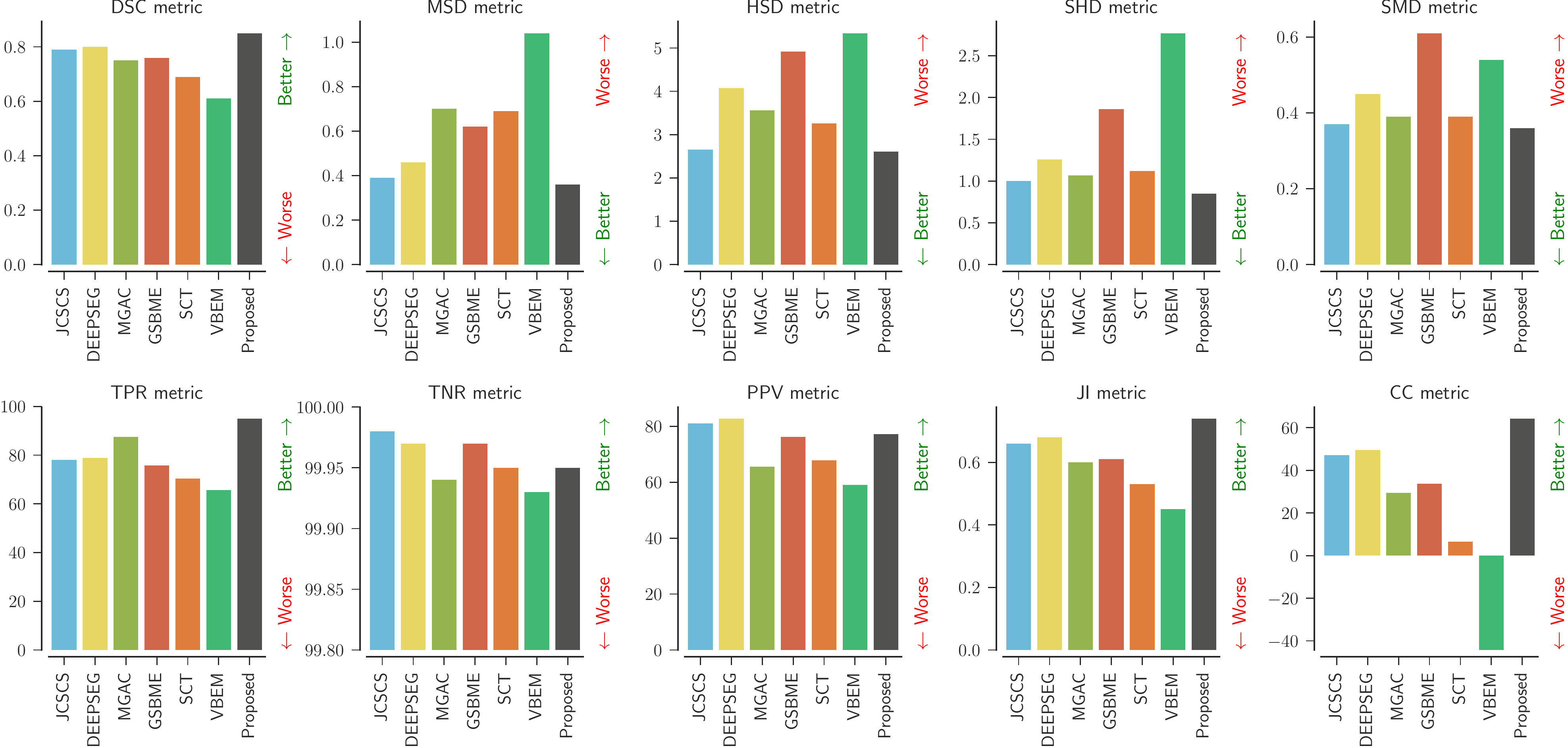}
	\caption{Test set evaluation results from the SCGM segmentation challenge \cite{Prados2017}  for each evaluated metric. Our method is shown as "Proposed". Best viewed in color.}
	\label{fig:challenge-results}
\end{figure*}

We can also see that the Dice Loss is not only an excellent surrogate for the Dice Similarity Coefficient (DSC) but also a surrogate for distance metrics, as can we note that our model not only achieved state-of-the-art results on overlap metrics (i.e. DSC) but also on distance and statistical metrics.

The True Negative Rate (TNR) and Positive Predictive Value (PPV) or precision, were the ones where the model didn't achieve the best results, however we note that the TNR was very close to the other methods results. We also hypothesize that the suboptimal results of the precision (PPV) are an effect of the sharp decision boundary produced by our model due to the Dice Loss. We believe that prediction threshold optimization can certainly yield better results, however this cost optimization would require further investigations.

When compared to Deepseg \cite{Porisky2017} method, the only method using Deep Learning in the challenge, where an U-Net based architecture was employed, 
our proposed approach performed better in 8 out of 10 metrics, even though our method didn't employed 3D convolutions, pre-training or threshold optimization as in Deepseg \cite{Porisky2017}. 

\begin{table*}[!ht]
	\centering
	\caption{Comparison of different segmentation methods that participated in the SCGM Segmentation Challenge \cite{Prados2017} against each of the four manual segmentation masks of the test set, reported here in the format: mean (std). For of fair comparison, the metrics are the same as used in \cite{Prados2017} and the results from other methods are replicated here, where we have: Dice similarity coefficient (DSC), mean surface distance (MSD), Hausdorff surface distance (HSD), skeletonized Hausdorff distance (SHD), skeletonized median distance (SMD), true positive rate (TPR), true negative rate (TNR), positive predictive value (PPV), Jaccard index (JI) and conformity coefficient (CC). In bold font, we represent the best-obtained results on each metric. We also note that MSD, HSD, SHD and SMD metrics are in millimeters and that lower values mean better results.}
	
	\label{tab:challenge-results}
	
	\resizebox{\textwidth}{!}{
		\begin{tabular}{@{}llllllll@{}}
			\toprule
			& \textbf{JCSCS}        & \textbf{DEEPSEG}      & \textbf{MGAC} & \textbf{GSBME} & \textbf{SCT} & \textbf{VBEM}  & \textbf{Proposed Method} \\ \midrule
			DSC & 0.79 (0.04)           & 0.80 (0.06)           & 0.75 (0.07)   & 0.76 (0.06)    & 0.69 (0.07)  & 0.61 (0.13)    & \textbf{0.85} (0.04)     \\
			MSD & 0.39 (0.44)           & 0.46 (0.48)           & 0.70 (0.79)   & 0.62 (0.64)    & 0.69 (0.76)  & 1.04 (1.14)    & \textbf{0.36} (0.34)     \\
			HSD & 2.65 (3.40)           & 4.07 (3.27)           & 3.56 (1.34)   & 4.92 (3.30)    & 3.26 (1.35)  & 5.34 (15.35)   & \textbf{2.61} (2.15)     \\
			SHD & 1.00 (0.35)           & 1.26 (0.65)           & 1.07 (0.37)   & 1.86 (0.85)    & 1.12 (0.41)  & 2.77 (8.10)    & \textbf{0.85} (0.32)     \\
			SMD & 0.37 (0.18)  & 0.45 (0.20)           & 0.39 (0.17)   & 0.61 (0.35)    & 0.39 (0.16)  & 0.54 (0.25)    & \textbf{0.36} (0.17)     \\
			TPR & 77.98 (4.88)          & 78.89 (10.33)         & 87.51 (6.65)  & 75.69 (8.08)   & 70.29 (6.76) & 65.66 (14.39)  & \textbf{94.97} (3.50)    \\
			TNR & \textbf{99.98} (0.03) & 99.97 (0.04)          & 99.94 (0.08)  & 99.97 (0.05)   & 99.95 (0.06) & 99.93 (0.09)   & 99.95 (0.06)             \\
			PPV & 81.06 (5.97)          & \textbf{82.78} (5.19) & 65.60 (9.01)  & 76.26 (7.41)   & 67.87 (8.62) & 59.07 (13.69)  & 77.29 (6.46)             \\
			JI  & 0.66 (0.05)           & 0.68 (0.08)           & 0.60 (0.08)   & 0.61 (0.08)    & 0.53 (0.08)  & 0.45 (0.13)    & \textbf{0.74} (0.06)     \\
			CC  & 47.17 (11.87)         & 49.52 (20.29)         & 29.36 (29.53) & 33.69 (24.23)  & 6.46 (30.59) & -44.25 (90.61) & \textbf{64.24} (10.83)   \\ \bottomrule
		\end{tabular}
	}
\end{table*}

\subsection{\textit{Ex vivo} high-resolution spinal cord}
In this subsection we show the evaluation on the \textit{ex vivo} high-resolution spinal cord data set.

\subsubsection{Qualitative Evaluation}
In the Figure \ref{fig:duke-preds}, we show a qualitative evaluation of the segmentations produced by our method and the U-Net model, contrasting the segmentations against the original and gold standard images.

As can be seen in the test sample depicted in the first column of Figure \ref{fig:duke-preds}, the predictions of the U-Net ``leaked'' the gray matter segmentation up to the cerebrospinal fluid (CSF) close to the dorsal horn (see green rectangle on first column), while our proposed segmentation was much more contained on the gray matter region only.

Also, in the third column of the Figure \ref{fig:duke-preds}, the U-Net significantly oversegmented a large portion of the gray matter region, extending the segmentation up to the white matter close to the right lateral horn of the gray matter anatomy (see the green rectangle), while our proposed method performed well.

\begin{figure*}
	\centering
	\includegraphics[width=\textwidth]{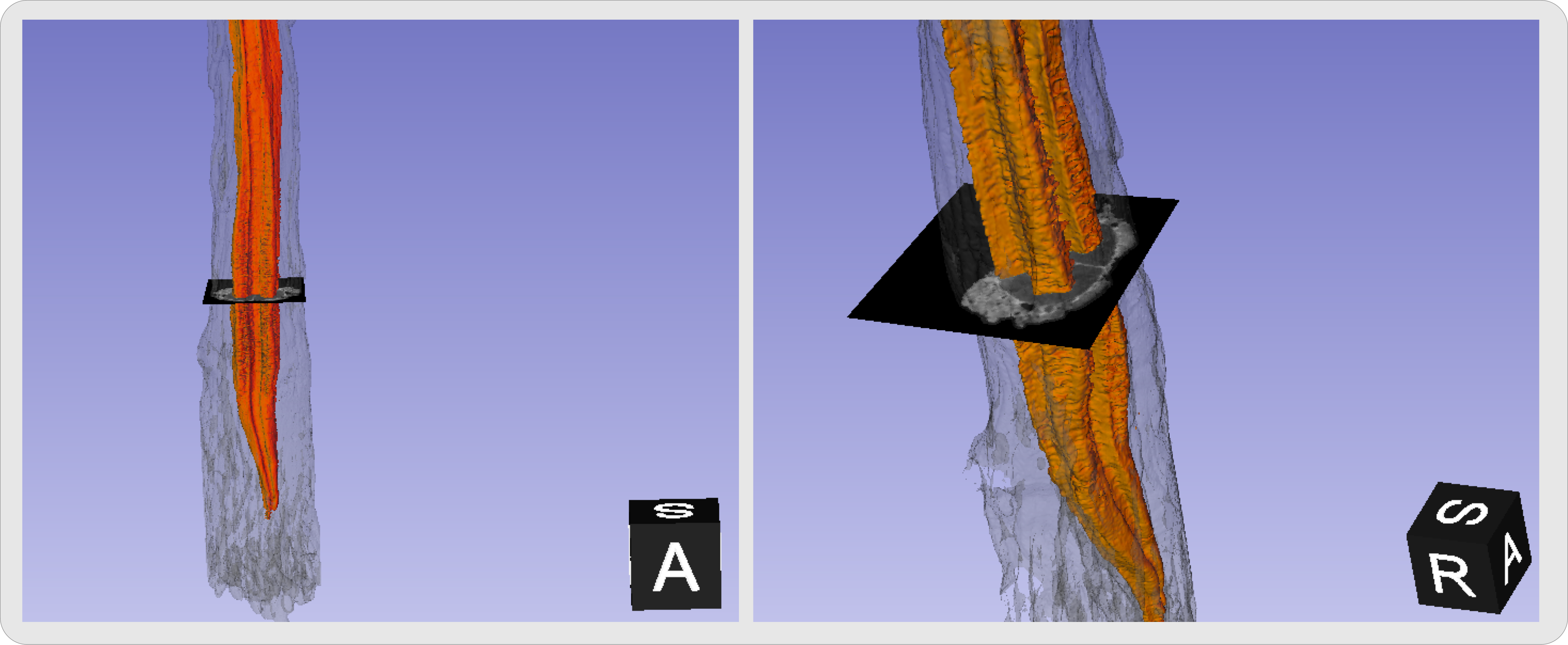}
	\caption{Lumbosacral region 3D rendered view of the \textit{ex vivo} high-resolution spinal cord data set segmented using the proposed method. The gray matter is depicted in orange color while the white matter and other tissues are represented in transparent gray color.}
	\label{fig:duke-3d-seg}
\end{figure*}

\begin{figure*}
	\centering
	\includegraphics[width=\textwidth]{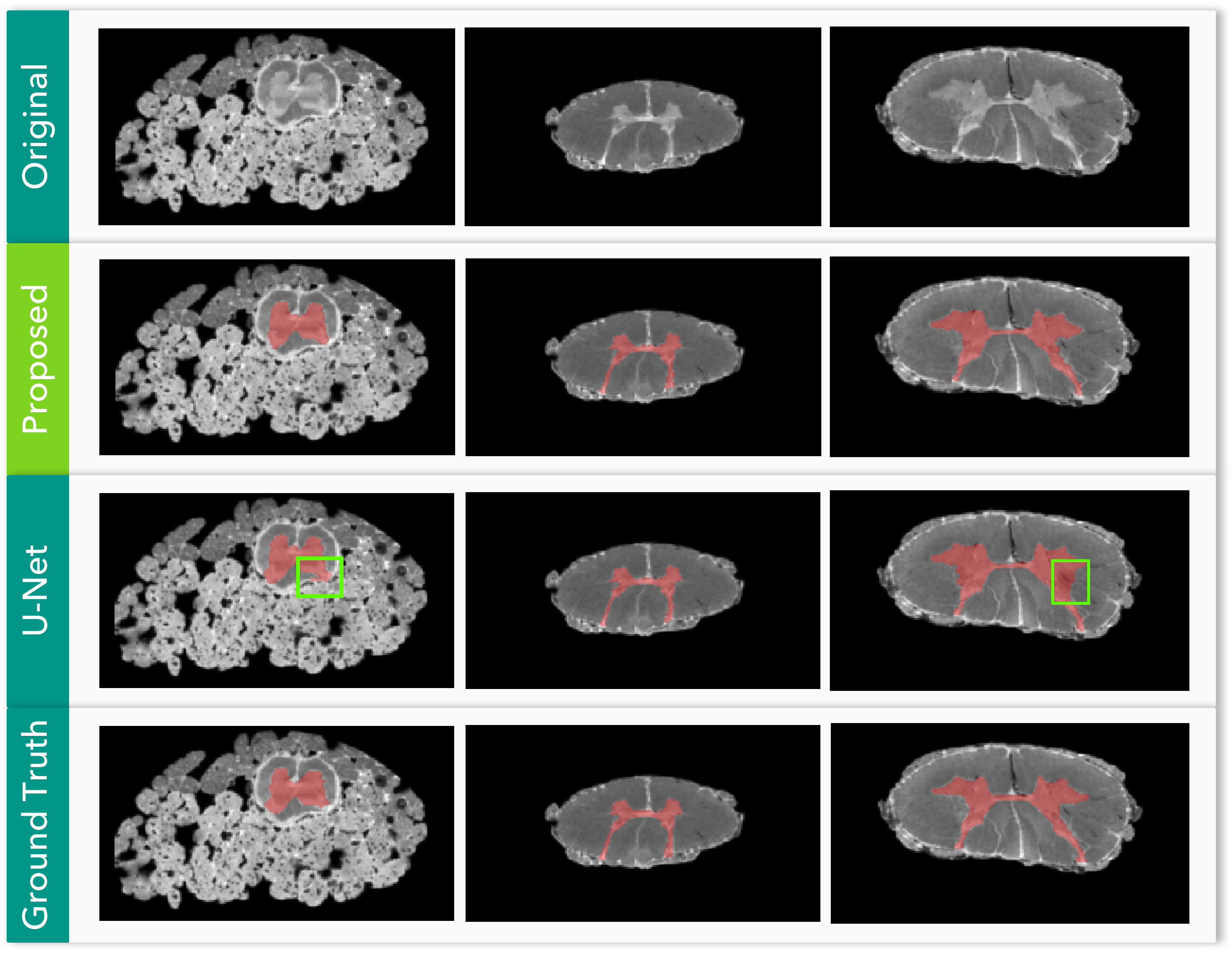}
	\caption{Qualitative evaluation of the U-Net and our proposed method on the \textit{ex vivo} high-resolution spinal cord data set. Each column represents a random sample of the test set (regions from left to right: sacral, thoracic, cervical). Green rectangles shows oversegmentation made by the U-Net model.}
	\label{fig:duke-preds}
\end{figure*}

We also provide in Figure \ref{fig:duke-3d-seg} a 3D rendered representation of the segmented gray matter using our method.

\subsubsection{Quantitative Evaluation}
As we can see in Table \ref{tab:duke}, where we show the quantitative results of our approach, our method achieved better results on 6 out of 8 different metrics. One of the main advantages that we can see from these results is that our method uses 6x less parameters than the U-Net architecture, leading to less chance of overfitting and potentially better generalization. 

During the training of the two architectures (U-Net and our method), we noticed that even with a high dropout rate of 0.4, the U-Net was still overfitting, forcing us to use a more aggressive data augmentation strategy to achieve better results, especially for the shifting parameters of the data augmentation; we hypothesize that this is an effect of the decimation on the contracting path of the U-Net, that disturbs the translational equivariance property of the network, leading to a poor performance on segmentation tasks.

\begin{table}[]
	\centering
	\caption{Quantitative metric results comparing a U-Net architecture and our proposed approach on the \textit{ex vivo} high-resolution spinal cord data set.}
	\label{tab:duke}
	\begin{tabular}{@{}lll@{}}
		\toprule
		\textbf{Metric name}         & \textbf{U-Net}       & \textbf{Proposed} \\ \midrule
		\textbf{Num. of Params.}  & 776,321              & \textbf{124,769}              \\
		\textbf{Dice}              & 0.9027 (0.07)          & \textbf{0.9226} (0.04)     \\
		\textbf{Mean Accuracy}     & \textbf{0.9626} (0.02) & 0.9561 (0.03)              \\
		\textbf{Pixel Accuracy}    & 0.9952 (0.01)          & \textbf{0.9968} (0.00)              \\
		\textbf{Recall}            & \textbf{0.9287} (0.05) & 0.9135 (0.06)              \\
		\textbf{Precision}         & 0.8831 (0.10)          & \textbf{0.9335} (0.04)     \\
		\textbf{Freq. Weighted IU} & 0.9913 (0.01)          & \textbf{0.9938} (0.00)     \\
		\textbf{Mean IU}           & 0.9121 (0.06)          & \textbf{0.9280} (0.04)     \\ \bottomrule
	\end{tabular}
\end{table}

\section{Discussion}
In this work, we devise a simple, but efficient and end-to-end method that achieves state-of-the-art results in many metrics when compared to six independently developed methods, as detailed in Table \ref{tab:challenge-results}. To the best of our knowledge, our approach is the first to achieve better results in 8 out of 10 metrics evaluated in the SCGM Segmentation Challenge \cite{Prados2017}.

One of the main differences with other methods from the challenge is that our method employs an end-to-end learning approach, where the entire prediction pipeline is optimized using backpropagation and gradient descent, contrasting with the other methods that usually employ separate registrations, external atlases/templates data and label fusion stages.

As we can also see in Table \ref{tab:duke}, when we compare our method to the most traditionally used method (U-Net) for medical image segmentation, our method provides not only better results in many metrics but also a major parameter reduction (more than 6 times). 

In the lens of Minimum Description Length (MDL) theory \cite{Rissanen1983}, that describes models as languages for describing properties of the data and sees inductive inference as finding regularity in the data \cite{Grunwald2007}, when two competing explanations for the data explains the data well, MDL will prefer the one that provides a shorter description of the data. As we can see, our approach using dilated filters provides more than 6x parameter reduction than U-Nets, but is also able to outperform other methods in many metrics, an evidence that the model is parameter-efficient and can capture a more compact description of the data regularities when compared with more complex models such as U-Nets.

Our approach is limited to 2D slices, however, the model doesn't restrict the use of 3D dilated convolutions and we believe that incorporating 3D context information into the model would certainly improve the segmentation results, however, at the expense of increased memory consumption.

We also believe that our method can be expanded to take leverage of semi-supervised learning approaches due to the strong smoothness assumption that holds for axial slices in most volumes, especially in \textit{ex vivo} high-resolution spinal cord MRI.

\section{Acknowledgments}
We acknowledge NVIDIA Corporation for the donation of a GPU Titan X board, Compute Canada for the GPU cluster, Zhuoqiong Ren for the help with gray matter gold standard, organizers of the SCGM Segmentation Challenge and participant teams that invested so much effort in this challenge. We also acknowledge United States National Institutes of Health awards P41 EB015897 and 1S10OD010683-01 for funding the \emph{ex vivo} study.

\section{Author Contributions}
C.S.P conceived the method, conducted the experiments, manual segmentations and wrote the paper. J.C.A. provided expert guidance and wrote the paper. E.C. provided the volume and information for the high-resolution \emph{ex vivo} dataset. All authors reviewed the paper.

\section{Additional Information}
\textbf{Competing financial interests}: C.S.P., E.C. and J.C.A. declare no competing financial interests.


\printbibliography[title={References}] 


\end{document}